\documentclass[10pt,twocolumn,letterpaper]{article}

\usepackage[pagenumbers]{cvpr} 

\usepackage{amsmath}
\usepackage{amssymb}
\usepackage{booktabs}
\usepackage{graphics}
\usepackage{graphicx}
\usepackage{float}
\usepackage{mathrsfs}
\usepackage{bbm}
\usepackage{capt-of,etoolbox}
\usepackage{multirow}
\usepackage{colortbl}
\usepackage{setspace}
\usepackage{stfloats}
\usepackage[colorlinks,linkcolor=red]{hyperref}
\newcommand{\cmmnt}[1]{}

\makeatletter
\@namedef{ver@everyshi.sty}{}
\makeatother

\usepackage[capitalize]{cleveref}
\crefname{section}{Sec.}{Secs.}
\Crefname{section}{Section}{Sections}
\Crefname{table}{Table}{Tables}
\crefname{table}{Tab.}{Tabs.}

\begin{document}

\title{UDE: A Unified Driving Engine for Human Motion Generation}


\author{
    Zixiang Zhou\\
    {\tt\small zhouzixiang@xiaobing.ai}\\
    \and
    Baoyuan Wang\\
    {\tt\small wangbaoyuan@xiaobing.ai}\\
}


\twocolumn[{
\maketitle
\begin{center}
    \includegraphics[width=0.95\textwidth]{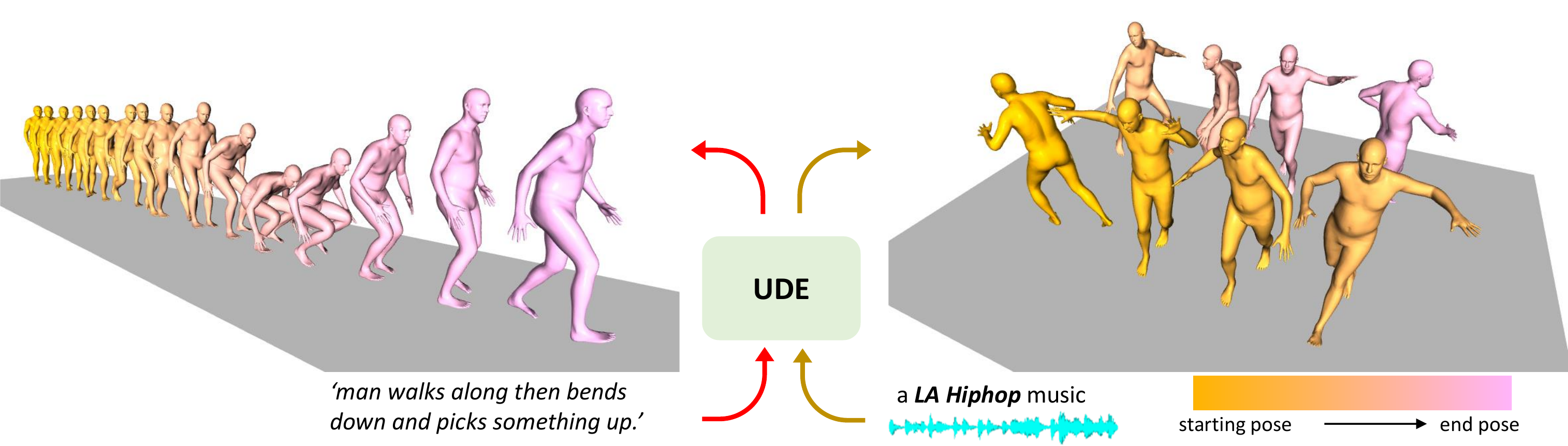}
    \captionof{figure}{Our shared Unified Driving Engine (\textbf{UDE}) can support both text-driven and audio-driven human motion generation. \textbf{Left} shows an example of a motion sequence driven by a text description while \textbf{Right} shows an example driven by a \textbf{\textit{LA Hiphop}} music clip.}
    \label{fig:teaser}
\end{center}
}]

\maketitle

\begin{abstract}
   Generating controllable and editable human motion sequences is a key challenge in 3D Avatar generation. It has been labor-intensive to generate and animate human motion for a long time until learning-based approaches have been developed and applied recently. However, these approaches are still task-specific or modality-specific\cite {ahuja2019language2pose}\cite{ghosh2021synthesis}\cite{ferreira2021learning}\cite{li2021ai}. In this paper, we propose ``UDE", the first unified driving engine that enables generating human motion sequences from natural language or audio sequences (see Fig.~\ref{fig:teaser}). Specifically, UDE consists of the following key components: 1) a motion quantization module based on VQVAE that represents continuous motion sequence as discrete latent code\cite{van2017neural}, 2) a modality-agnostic transformer encoder\cite{vaswani2017attention} that learns to map modality-aware driving signals to a joint space, and 3) a unified token transformer (GPT-like\cite{radford2019language}) network to predict the quantized latent code index in an auto-regressive manner. 4) a diffusion motion decoder that takes as input the motion tokens and decodes them into motion sequences with high diversity. We evaluate our method on HumanML3D\cite{Guo_2022_CVPR} and AIST++\cite{li2021learn} benchmarks, and the experiment results demonstrate our method achieves state-of-the-art performance. Project website: \url{https://github.com/zixiangzhou916/UDE/}
\end{abstract}

\section{Introduction}

\setlength{\parindent}{1em} Synthesizing realistic human motion sequences has been a pillar component in many real-world applications, such as the game and film industry, robot control, etc. It is labor-intensive and tedious work to design and synthesize human motion sequences from scratch using professional software. It requires professional skills in human kinematics, animation creation, and software operation to achieve the creation of one single piece of motion sequence synthesis, making it hard to be democratized for broad content generations. 
Recently, the emergence of motion capture and pose estimation \cite{joo2018total}\cite{zhang2021lightweight}\cite{rong2020frankmocap}\cite{yuan2022glamr} have made it possible to synthesize human motion sequences from VTubers or source videos thanks to the advances of deep learning. Although these approaches have simplified the creation of motion sequences, actors or highly correlated videos are still necessary, thus limiting the scalability as well as the controllability.\par

\setlength{\parindent}{1em} The development of multi-modal machine learning paves a new way to human motion synthesis\cite{ahuja2019language2pose}\cite{ghosh2021synthesis}\cite{Guo_2022_CVPR}\cite{huang2020dance}\cite{li2022danceformer}\cite{chen2021choreomaster}. For example, natural language descriptions could be used to drive human motion sequences  directly\cite{ahuja2019language2pose}\cite{ghosh2021synthesis}\cite{Guo_2022_CVPR}. The language description is a straightforward representation for human users to control the synthesis. It provides a semantic clue of what the synthesized motion sequence should look like, and the editing could be conducted by simply changing the language description. Language, however, does not cover the full domain of human motion sequences. In terms of dancing motion synthesis, for example, the natural language is not sufficient to describe the dance rhythm. For such scenarios, audio sequences are used as guidance to help motion synthesis. The dance rhythm and choreography style are expressed by the audio sequence, so the synthesized motion could match the music beat rhythmically and choreography style. However, these approaches are studied separately in prior works. In many real-world applications, the characters are likely to perform a complex motion sequence composed of both rhythmic dances from music and certain actions described by language. As a result, multi-modal motion consistency would become an urgent issue to solve if employed siloed modality-specific models.\par

\setlength{\parindent}{1em} To address above mentioned problems, in this work, we propose a \textbf{U}nified \textbf{D}riving \textbf{E}ngine (\textbf{UDE}) which unifies the human motion generation driven by natural language and music clip in one shared model. Our model consists of four key components. First, we train a codebook using VQ-VAE. For the codebook, each code represents a certain pattern of the motion sequence. Second, we introduce a Modality-Agnostic Transformer Encoder (\textbf{MATE}). It takes the input of different modalities and transforms them into sequential embedding in one joint space. The third component is a Unified Token Transformer (\textbf{UTT}). We feed it with sequential embedding obtained by \textbf{MATE} and predict the motion token sequences in an auto-regressive manner. The fourth component is a Diffusion Motion Decoder (\textbf{DMD}). Unlike recent works\cite{tevet2022human}\cite{zhang2022motiondiffuse}, which are modality-specific, our $\textbf{DMD}$ is modality-agnostic. Given the motion token sequences, \textbf{DMD} encodes them to semantic-rich embedding and then decodes them to motion sequences in continuous space by the reversed diffusion process.\par

\setlength{\parindent}{1em} We summarize our contributions in four folds: \textbf{1}) We model the continuous human motion generation problem as a discrete token prediction problem. By learning a context-rich codebook, we can generate long motion sequences with high motion quality and semantic consistency. \textbf{2}) We unify the text-driven and audio-driven motion generation into one single unified model. By learning \textbf{MATE}, we can map input sequences of different modalities into joint space. Then we can predict motion tokens with \textbf{UTT} regardless of the modality of input. \textbf{3}) We propose \textbf{DMD} to decode the motion tokens to motion sequence. Compared to the decoder in VQ-VAE, which generates deterministic samples, our \textbf{DMD} can generate samples with high diversity. \textbf{4}) We evaluate our method extensively and the results suggest that our method outperforms existing methods in both text-driven and audio-driven scenarios.\par
\vspace{-3mm}

\section{Related Work}

\paragraph{Text to Motion}The recent success of multi-modal machine learning makes it possible to synthesize human motion from text descriptions. \cite{ahuja2019language2pose} proposed a method to generate human motion from natural language. They learn joint embedding between language and poses with different encoders, and use a GRU-based motion decoder to map the embedding to human motion. \cite{ghosh2021synthesis} further proposed to learn a joint embedding among natural language, human upper body, and lower body. They use a two-stream encoder to map the upper body poses and lower body poses to the joint embedding space, and a pre-trained BERT model\cite{devlin2018bert} to encode the text description. In general, multiple motion sequences could be derived from a single text description. To enable probabilistic text-guided synthesis, \cite{petrovich2022temos} proposed to learn a joint distribution from motion sequence and natural language. Instead of learning a continuous latent space, discrete latent space is also successfully verified in representing human motion\cite{guo2022tm2t}. In this work, motion sequences are encoded and discretized into a codebook. A text-to-motion module is proposed to predict motion code index from text descriptions, and a motion-to-text module is used to introduce cycle consistency. In addition, motion synthesis also benefits from the recent development of multi-modal pre-training such as CLIP\cite{radford2021learning}. The concept of CLIP is employed in \cite{tevet2022motionclip} to synthesize human motion from natural language. In this work, they try to minimize the distance between the embeddings of paired motion and text. Recently, Diffusion models have emerged as an alternative in motion generation\cite{tevet2022human}\cite{zhang2022motiondiffuse}. They encode text descriptions using pretrained models, and estimate the Gaussian noise\cite{zhang2022motiondiffuse} and final results\cite{tevet2022human} directly at every reversed diffusion step.
\vspace{-3mm}
\paragraph{Music to Motion}Music-to-motion, compared with text-to-motion, has different philosophy. There is no strict mapping between music and motion, but the rhythm, beat, and style\cite{chen2021choreomaster} are critical points that build the correlation between music and motion. \cite{ferreira2021learning} proposed a GAN-based method to synthesize dance motion from music input. They use a CNN encoder to extract music features and use an ST-GCN\cite{yan2018spatial} module to decode the dance motions. Gaussian noise is added to take diversity into consideration. An auto-regressive approach is introduced in \cite{huang2020dance}, where the music sequence is encoded by a transformer encoder at first. Then an RNN-based decoder is employed to predict step-wise pose given music feature and previous dance poses, where a Begin of Pose is used as an initial. Similarly, \cite{ren2019music}\cite{li2021ai} synthesize dance poses auto-regressively, but with transformer\cite{vaswani2017attention} architectures. These approaches encode audio sequence and previous motion sequence with different transformer encoders, respectively, and a fusion module is used to synthesize future poses conditioned on music and dance poses. These two approaches are deterministic, meaning that the same condition always gets the same results. To introduce diversity, \cite{valle2021transflower} proposed a normalizing-flow\cite{rezende2015variational} based approach. In this work, the same cross-modal feature described in \cite{li2021ai} is used to extract the conditioned feature, while a normalizing-flow module\cite{henter2020moglow} is used to generate poses given the cross-modal feature and random noise. \cite{kim2022brand} proposed a GAN-based approach to synthesize diverse dance motions conditioned on music input. They use a transformer decoder to map the input music and seed poses to long-range motion poses, where the dance genre is constrained by mapping the genre to a style latent code. A transformer encoder is used as a discriminator to distinguish the synthesized motion between fake and real. Instead of representing the condition information with continuous latent, \cite{siyao2022bailando} proposed to use a discrete representation. Their approach first learns a codebook using VQ-VAE, then a GPT-like transformer module is used to predict the motion code index given music input. In addition, they proposed to decompose a human pose into its upper body and lower body parts, and the final synthesized pose is a composition of upper body and lower body poses.

\setlength{\parindent}{1em}These methods, however, are all modality-specific. There is a lack of a solution to unify the multi-modality driven human motion generation tasks.

\section{Method}
\begin{figure*}[ht]
    \centering
    \includegraphics[width=\textwidth]{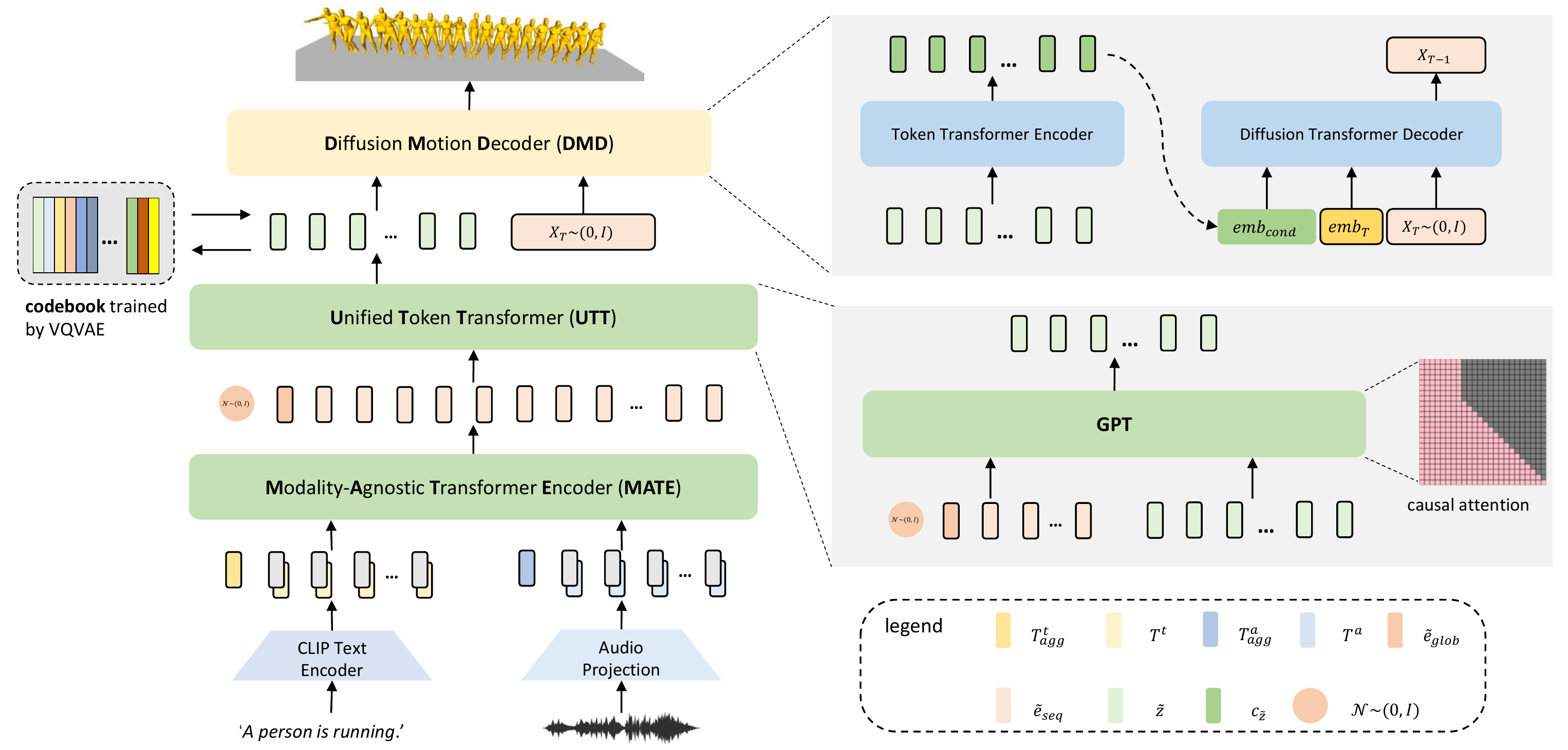}
    \caption{\textbf{Pipeline of our method.} \textbf{Left} is the overview of our method. We learn a codebook by training a VQVAE we called Motion Quantization (\textbf{\textit{MQ}}). Then a Modality-Agnostic Transformer Encoder (\textbf{\textit{MATE}}) takes as input text description or audio sequence and maps them to sequential embedding $\tilde{e}$ in joint space. The sequential embedding $\tilde{e}$ is then fed to Unified Token Transformer Decoder ($\textbf{\textit{UTT}}$) to predict the token sequence $\tilde{z}$. Finally, we propose a Diffusion Motion Decoder ($\textbf{\textit{DMD}}$) to map the token sequence $\tilde{z}$ to a motion sequence with high diversity. The details of $\textbf{\textit{UTT}}$ and $\textbf{\textit{DMD}}$ are shown in the zoomed-in panels on the $\textbf{Right}$.}
    \label{fig:overview}
\end{figure*}

The overview of our entire framework is illustrated in Fig. \ref{fig:overview}, which contains four modules: \textbf{1}) Motion Quantization module, \textbf{2}) Modality-Agnostic Transformer module, \textbf{3}) Unified Token Transformer module, and \textbf{4}) Diffusion Motion Decoder module, which will be described in the following respectively.

\subsection{Motion Quantization (\textbf{MQ}) }
We learn a semantic-rich codebook by training a VQ-VAE model. We denote a motion sequence as $x \in \mathbbm{R}^{T\times c}$, where \textit{T} is the length of the motion sequence, and \textit{c} is the dimension per frame. To learn the codebook $\mathcal{Z}=\{z^q|z^q\in\mathbbm{R}^{T^{q}\times d}\}$, we train an encoder to map $\textit{x}$ to $\textit{e}\in\mathbbm{R}^{T'\times d}$.
Given the sequential embedding $\textit{e}\in\mathbbm{R}^{T'\times d}$, we quantize them by replacing each \textit{$e_i$} with its nearest code \textit{$z^q_j$} in $\mathcal Z$ as Eq. \eqref{eq:quantization}
\begin{equation}
    \mathcal{Q}({e_i})=\mathop{\arg\min}\limits_{z^q_j \in \mathcal{Z}}\|e_i - z^q_j\| \label{eq:quantization}
\end{equation}
To reconstruct the motion from $\textit{z}^q$, we employ a decoder to decode the sequential codes $\textit{z}^q \in \mathbbm{R}^{T'\times d}$ back to motion sequence $\tilde{x} = \mathcal{D}(z^q)\in\mathbbm{R}^{T'\times d}$. 
The encoder, decoder, and codebook are trained simultaneously by optimizing the following loss function:
\begin{equation}
    \mathcal{L}_{VQ} = \mathcal{L}_{rec}(\tilde{x}, \textit{x}) + \beta_1 \|sg[\textit{e}] - \mathcal{Q}(e)\| + \beta_2 \|\textit{e} - sg[\mathcal{Q}(e)]\|
    \label{eq:loss_vqvae}
\end{equation}
\noindent In Eq.\eqref{eq:loss_vqvae}, $\mathcal{L}_{rec}(\tilde{x}, \textit{x})$ is the reconstruction term, which encourages the decoded $\tilde{x}$ to be close to \textit{x} as much as possible. The second term, $\|sg[\textit{e}] - \mathcal{Q}(e)\|$, is the codebook loss, which encourages the $\textit{z}^q_j$ to move close to encoded embedding $\textit{e}_i$. The third term $\|\textit{e} - sg[\mathcal{Q}(e)]\|$ is commitment loss, it encourages the embedding $\textit{e}_i$ to stay close to corresponding discrete codes $\textit{z}^q_j$ so that the training process could be stabilized. Both the encoder and decoder adopt a 1D temporal convolution architecture.

\subsection{Motion-Agnostic Transformer Encoder(\textbf{MATE})}
\setlength{\parindent}{1em} \textbf{MATE} is designed to convert multi-modal input data to modality-agnostic output. Our encoder takes as input two different modalities, text descriptions, and audio sequences. For text input, the CLIP text encoder\cite{radford2021learning} is used to extract word-level embedding. We skip the last step in CLIP text encoder, which is a max-pooling operation, to obtain the word level sequential embedding of input text description as $\tilde{e}^t = \mathcal{E}(\textit{x}_t)$. For audio input, we simply apply a linear layer to project the raw audio input sequential feature vectors as $\tilde{e}^a = \mathcal{E}(\textit{x}_a)$. For simplicity, we express our \textbf{MATE} mapping as $\tilde{e}^k = \mathcal{E}(\textit{x}^k)$, where $\textit{x}^k$ stands for the input of modality $\textit{k}$, here $\textit{k}$ could either refer to $\textit{audio}$ or $\textit{text}$. Before feeding sequential feature vectors of each modality into the model, we add learnable token embedding for each modality to them and also prepend learnable aggregation token embedding to the feature sequence. Finally, we apply position encoding to the feature sequence to obtain the final input sequence:

\begin{equation}
    \tilde{e} = [\textit{T}_{agg}^k, \tilde{e}_0^k + \textit{T}^k, \tilde{e}_1^k + \textit{T}^k, ..., \tilde{e}_i^k + \textit{T}^k, ..., \tilde{e}_I^k + \textit{T}^k] + \textit{e}_{pos}
    \label{eq:add_tokens}
\end{equation}

\noindent where in Eq \eqref{eq:add_tokens}, $\textit{T}_{agg}^k$ is the learnable aggregation token embedding for modality $\textit{k}$, and $\textit{T}_k$ is the learnable token embedding added to feature sequence of modality $\textit{k}$. The transformer architecture follows \cite{dosovitskiy2020image}, where $\textit{n}$ transformer encoder layers are stacked and full self-attention mechanism is employed. For the output sequence of length $\textit{m}$, we take the first element as the global embedding $\tilde{e}_{glob}$, while the rest $\textit{m}-1$ elements are sequential embedding $\tilde{e}_{seq}$, respectively.\par

\subsection{Unified Token Transformer (\textbf{UTT})}
\setlength{\parindent}{1em}Our \textbf{UTT} adopts a stacked transformer encoder layers architecture with a causal attention mechanism. We feed the embedding $\tilde{e}_{glob}$ and $\tilde{e}_{seq}$ to \textbf{UTT} as conditions, as well as the embedding of target motion tokens $\tilde{e}_{mot}$. We concatenate them along temporal dimension as $\textit{e}_{in} = [\tilde{e}_{glob}, \tilde{e}_{seq}, \tilde{e}_{mot}]$. We employ causal self-attention in training the \textbf{UTT} to make sure the future information is inaccessible. However, we want the condition information always be accessible during the training process, we don't mask the region of conditions, but only mask out the region corresponding to future motion tokens. Denote the condition as $\tilde{e}_{cond}^{0:L}$ and motion tokens embedding as $\tilde{e}_{mot}^{0:T}$, the attention region of motion token embedding $\tilde{e}_{mot}^{i}$ is $[\tilde{e}_{cond}^{0:L}, \tilde{e}_{mot}^{<=i}]$. Then $\textbf{UTT}$ transforms $\textit{e}_{in}$ to token sequence $\textit{z}^q\in\mathbbm{R}^{T\times d}$ auto-regressively.\par

\setlength{\parindent}{1em} As shown in Fig. \ref{fig:overview}, we can inject $\textit{z}\thicksim\mathcal{N}(0,I)$ for diversity. Given a sampled $\textit{z}$, we map it to $\tilde{z}$ through a MLP, so that $\textit{z}$ has the same dimension as $\tilde{e}_{glob}$, then we get the new global condition embedding as $\tilde{e}_{glob}^{\tilde{z}}=\tilde{z} + \tilde{e}_{glob}$. So the new input to \textbf{UTT} now becomes $e_{in}=[\tilde{e}_{glob}^{\tilde{z}}, \tilde{e}_{seq}, \tilde{e}_{mot}]$.\par

\setlength{\parindent}{1em} We introduce a discriminator to help training the $\textbf{UTT}$ and $\textbf{MATE}$ end-to-end. We use a conditional discriminator, which takes as input the global embedding $\tilde{e}_{glob}$ and the motion sequence. Instead of predicting one score per sequence as conventional discriminators do, we adopt the strategy described in PatchGAN\cite{isola2017image}. The motion sequence $\textit{x}\in\mathbbm{R}^{T\times c}$ is fed to the discriminator and the sequential feature vector $\tilde{e}_{\textit{x}}^D \in \mathbbm{R}^{T'\times c}$ is extracted by a 1D temporal convolution architecture, where $\textit{T}' = T / 4$. We apply a linear layer to the $\tilde{e}_{glob}$ to get $\tilde{e}_{glob}^D=FC(\tilde{e}_{glob})$ so that both $\tilde{e}_{\textit{x}}^D$ and $\tilde{e}_{glob}^D$ have same dimension. Then we feed the added features $\tilde{e}^D=\tilde{e}_{glob}^D + \tilde{e}_{\textit{x}}^D$ to a two-layer transformer encoder to compute its sequential embedding. For each embedding, a linear projection is applied to transform it into a validity score.

The overall objective of $\textbf{MATE}$ and $\textbf{UTT}$ is:
\vspace{-1.5mm}
\begin{equation}
    \mathcal{L} = \mathcal{L}_{ce} + \beta_{adv}\mathcal{L}_{adv}
    \label{eq:over-loss}
\end{equation}
\vspace{-0.5mm}
\noindent where $\mathcal{L}_{ce}$ is the cross entropy loss to token prediction, and $\mathcal{L}_{adv}$ is the adversarial loss on motion sequence.
\begin{equation}
    \mathcal{L}_{ce} = -\sum_{i=1}^{I}\textit{p}(\textit{z}^q_i)\log\textit{q}_\theta(\tilde{z}^q_i)
    \label{eq:ce-loss}
\end{equation}
\begin{equation}
    \mathcal{L}_{adv} = -\mathbbm{E}_{p \thicksim \textit{P}_{gen}, q \thicksim \textit{Q}_{data}}\textit{D}(\textit{G}(\textit{p}), \textit{q})
    \label{eq:adv-loss}
\end{equation}
\noindent where in Eq \eqref{eq:adv-loss}, $\textit{p} \thicksim \textit{P}_{gen}$ is the generated sample distribution, and $\textit{q} \thicksim \textit{Q}_{data}$ is the real sample distribution, respectively. $\beta_{adv}$ is the balancing weight.\par

\subsection{Diffusion Motion Decoder (\textbf{DMD})}
\setlength{\parindent}{1em}The pre-trained VQ-decoder produces deterministic outputs given same input tokens $\tilde{x} = \mathcal{D}(\textit{z}^q)$. However, diversity is also desirable at the token decoding stage. We propose a diffusion motion decoder to replace the VQ-decoder to introduce additional diversity. Unlike \cite{tevet2022human}\cite{zhang2022motiondiffuse}, which take as input text descriptions directly, making them modality-specific. Our method is a modality-agnostic model which takes as input discrete tokens as condition, regardless what modality the raw input is. The diffusion process\cite{ho2020denoising} is a Markov noising process, starting from real data $\textit{X}_0$, Gaussian noise is added at each step to convert $\textit{X}_0$ to $\textit{X}_{T} \thicksim \mathcal{N}(0, \textit{I})$. This process is expressed as
\vspace{-3mm}

\begin{equation}
    \textit{q}(\textit{X}_{t}|\textit{X}_{t-1}) = \mathcal{N}(\textit{X}_t; \sqrt{1 - \beta_{t}}\textit{X}_{t-1}, \beta\textbf{I})
    \label{eq:one-step-forward-process}
\end{equation}

\noindent The entire diffusion process could be formulated as
\vspace{-3mm}

\begin{equation}
    \textit{q}(\textit{X}_{1:\textit{T}} | \textit{X}_0) = \prod \limits_{t=0}^T\textit{q}(\textit{X}_t | \textit{X}_{t-1})
    \label{eq:forward-process}
\end{equation}

\noindent Letting $\alpha_t = 1 - \beta_t$ and $\tilde{\alpha}_t = \prod \limits_{i=1}^t\alpha_i$, the noisy data at arbitrary step $\textit{t}$ could be derived from $\textit{X}_0$ as 

\begin{equation}
    \textit{q}(\textit{X}_{t} | \textit{X}_0) = \mathcal{N}(\textit{X}_t; \sqrt{\tilde{\alpha}_t}\textit{X}_0, (1 - \tilde{\alpha}_t)\textbf{I})
    \label{eq:perfect-property}
\end{equation}

\noindent The reversed diffusion process attempts to gradually denoise $\textit{X}_t$. In our context, this reversed diffusion process is conditioned on $\textit{c}_{\tilde{z}} = \mathcal{E}_{z}(\textit{z}^q)$, where $\textit{z}^q$ is the predicted tokens in the codebook, and $\mathcal{E}_z(\textit{z}^q)$ extracts embedding from the token sequence. We follow the strategy in \cite{ho2020denoising} where we predict the noise $\epsilon$ added to $\textit{X}_t$ as $\epsilon_t = \epsilon_\theta(\textit{X}_t, \textit{t}, \textit{c}_{\tilde{z}})$. Our diffusion motion decoder is shown in Fig.~\ref{fig:overview}. It consists of two parts. The first part is a token transformer encoder, which maps codebook token sequence to sequential embedding as $\textit{c}_{\tilde{z}} = \mathcal{E}_{z}(\textit{z}^q)$. The sequential embedding $\textit{c}_{\tilde{z}}$ is served as condition embedding. For each reversed diffusion step $\textit{t}$, the diffusion transformer decoder takes as input the sequential embedding $\textit{c}_{\tilde{z}}$, the embedding of timestep $\textit{emb}_T$, and the noisy data $\textit{X}_t$, and predicts the noise as $\epsilon_t = \epsilon_{\theta}(\textit{c}_{\tilde{z}}, \textit{emb}_{T}, \textit{X}_t)$, where $\theta$ are learnable parameters of the diffusion transformer decoder. We train the diffusion motion decoder by optimizing the following objective as
\begin{equation}
    \mathcal{L}_{diff} = \mathbbm{E}_{t \in [1, T], \textit{X}_0 \thicksim \textit{q}(\textit{X}_0), \epsilon \thicksim \mathcal{N}(0, \textbf{I})}[\|\epsilon - \epsilon_{\theta}(\textit{c}_{\tilde{z}}, \textit{emb}_{T}, \textit{X}_t)\|]
    \label{eq:}
\end{equation}

\section{Experiments}
We implement our method from end2end and evaluate it on two types of tasks: text-to-motion generation and audio-to-motion generation, which will be described in detail.\par

\begin{figure*}[ht]
    \centering
    \includegraphics[width=\linewidth]{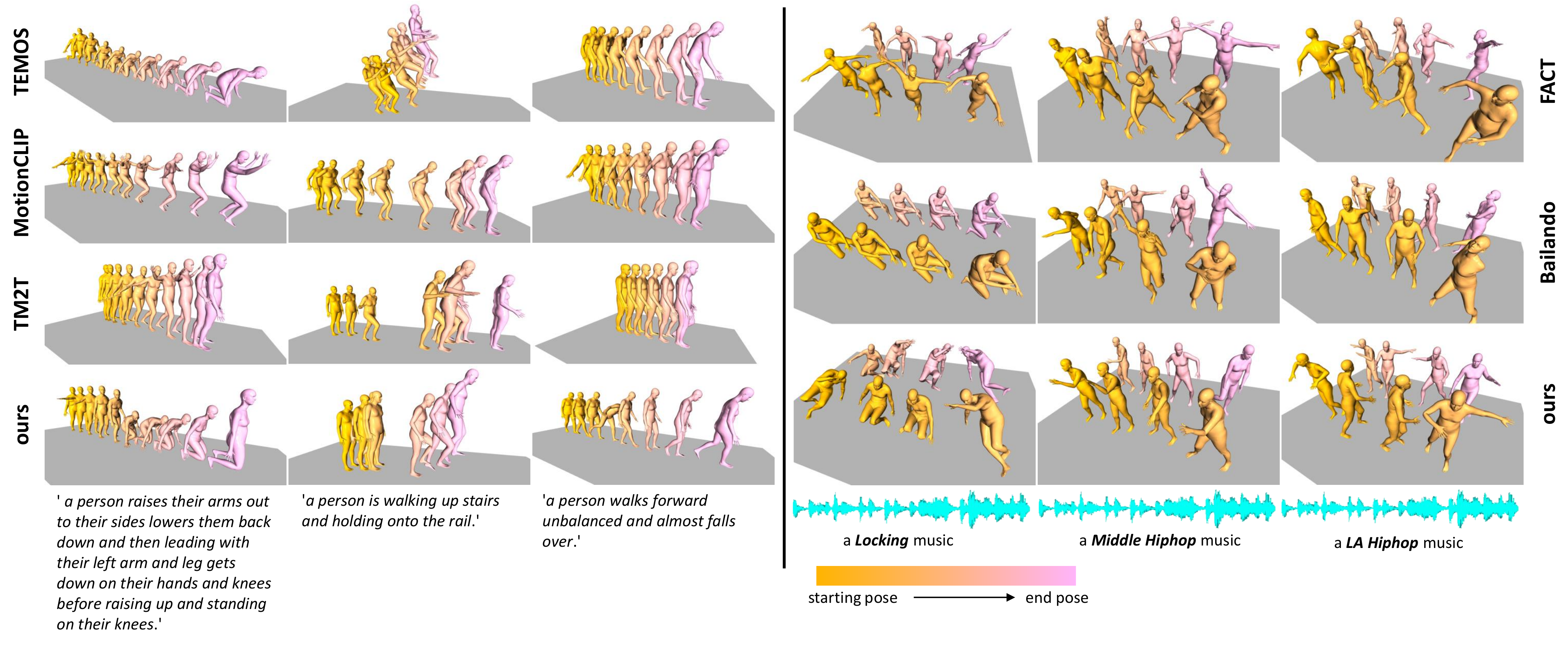}
    \caption{\textbf{Qualitative comparison with existing methods} on Text-to-Motion (\textbf{Left}) and Audio-to-Motion (\textbf{Right}) tasks. \textbf{Left} shows the comparison between ours and \cite{petrovich2022temos}\cite{tevet2022motionclip}\cite{guo2022tm2t} driven by the same text descriptions. \textbf{Right} shows the comparison between ours and \cite{li2021ai}\cite{siyao2022bailando}. We appropriately adjust the position of each pose for better visualization.}
    \label{fig:compare}
\end{figure*}

\begin{table*}[]
    \centering
    \resizebox{0.9\linewidth}{!}{
        \begin{tabular}{ccccccccccc}
            \hline
            \multirow{2}{*}{Method} & \multicolumn{2}{c}{Text Retrieval Acc.} & \multicolumn{2}{c}{FID} & \multicolumn{2}{c}{Diversity} & \multicolumn{4}{c}{Recon Acc.} \\
            \cline{2-11}
             & Top-1 Acc. $\uparrow$ & Top-5 Acc. $\uparrow$ & $\rm FID_k$ $\downarrow$ & $\rm FID_m$ $\downarrow$ & $\rm Div_k$ $\uparrow$ & $\rm Div_m$ $\uparrow$ & APE $\downarrow$ & AVE $\downarrow$ & APE(root) $\downarrow$ & AVE(root) $\downarrow$ \\
            \hline
            GT & 9.99 & 27.74 & 4.84 & 1.11 & 7.82 & 6.47 & 0.00 & 0.00 & 0.00 & 0.00 \\
            \hline
            TEMOS\cite{petrovich2022temos} & 4.86 & 16.58 & 45.31 & 21.37 & 3.48 & 7.63 & 0.24 & 0.03 & 0.50 & 0.38 \\
            MotionCLIP \cite{tevet2022motionclip} & 7.01 & 22.49 & 37.46 & 10.76 & 4.11 & \textbf{9.07} & 0.26 & 0.03 & 0.52 & 0.37 \\
            TM2T \cite{guo2022tm2t} & \underline{7.76} & \underline{22.54} & \textbf{13.25} & \underline{6.39} & \underline{4.78} & 6.34 & \textbf{0.19} & \textbf{0.02} & \textbf{0.44} & \underline{0.34} \\
            \hline
            \rowcolor[rgb]{0.902,0.898,0.898} Ours & \textbf{8.21} & \textbf{26.06} & \underline{19.35} & \textbf{2.67} & \textbf{6.75} & \underline{7.82} & \textbf{0.19} & \textbf{0.02} & \textbf{0.44} & \textbf{0.33} \\
            \hline
        \end{tabular}
    }
    \caption{\textbf{Quantitative results of text-to-motion task under various metrics.} For fair comparison, we reproduce the results of \cite{petrovich2022temos}\cite{tevet2022motionclip}\cite{guo2022tm2t} on HumanML3D dataset \cite{Guo_2022_CVPR} using same splits. The motion representation is the same as ours and trained using their official codes. For ours, we generate 3 samples per input with Unified Token Transformer (\textbf{\textit{UTT}}), and for each token sequence, we generate 30 samples with Diffusion Motion Decoder (\textbf{\textit{DMD}}). Then take the average metrics. \textbf{Bold} indicates best results, and \underline{underlined} indicates the second best.}
    \label{tab:text-to-motion-table}
\end{table*}

\begin{table*}[ht]
    \centering
    \resizebox{0.8\linewidth}{!}{
        \begin{tabular}{cccccccccc}
             \hline
            \multirow{2}{*}{Method} & \multirow{2}{*}{Beat Align $\uparrow$} & \multicolumn{2}{c}{FID} & \multicolumn{2}{c}{Diversity} & \multicolumn{4}{c}{Recon Acc.} \\
             \cline{3-10}
            & & $\rm FID_k$ $\downarrow$ & $\rm FID_m$ $\downarrow$ & $\rm Div_k$ $\uparrow$ & $\rm Div_m$ $\uparrow$ & APE $\downarrow$ & AVE $\downarrow$ & APE(root) $\downarrow$ & AVE(root) $\downarrow$ \\
             \hline
             GT & 0.237 & 17.10 & 10.60 & 8.19 & 7.45 & 0.00 & 0.00 & 0.00 & 0.00 \\
             \hline
             FACT \cite{li2021ai} & 0.2209 & 35.35 & 22.11 & 5.94 & \underline{6.18} & \textbf{0.26} & 0.04 & 0.70 & \textbf{0.11} \\
             Bailando \cite{siyao2022bailando} & \textbf{0.2332} & \underline{28.16} & \underline{9.62} & \textbf{7.83} & \textbf{6.34} & \underline{0.27} & \textbf{0.03} & \underline{0.32} & 0.13 \\
             \hline
             \rowcolor[rgb]{0.902,0.898,0.898} Ours & \underline{0.2311} & \textbf{17.25} & \textbf{8.69} & \underline{7.78} & 5.81 & 0.29 & \textbf{0.03} & \textbf{0.28} & \underline{0.12} \\
             \hline
        \end{tabular}
    }
    \caption{\textbf{Quantitative results of audio-to-motion task.} For ours, we generate 3 samples per input with Unified Token Transformer (\textbf{\textit{UTT}}), and for each token sequence, we generate 30 samples with Diffusion Motion Decoder (\textbf{\textit{DMD}}). Then we take the average metrics.}
    \label{tab:audio-to-motion-table}
\end{table*}

\begin{table}[ht]
    \centering
    \resizebox{\linewidth}{!}{
        \begin{tabular}{ccc|cccc}
            \hline
            \multirow{2}{*}{Method} & \multicolumn{2}{c|}{Text-to-Motion} & \multicolumn{4}{c}{Audio-to-Motion} \\
            \cline{2-7}
             & Top-1 Acc. $\uparrow$ & Top-5 Acc. $\uparrow$ & $\rm FID_K$ $\downarrow$ & $\rm FID_m$ $\downarrow$ & $\rm Div_k$ $\uparrow$ & $\rm Div_m$ $\uparrow$ \\
            \hline
            Ours (gru) & 7.89 & 23.45 & 47.21 & 23.13 & \textbf{6.71} & 3.72 \\
        \rowcolor[rgb]{0.902,0.898,0.898} Ours (gpt) & \textbf{8.11} & \textbf{25.01} & \textbf{28.44} & \textbf{15.70} & 6.13 & \textbf{4.07} \\
            \hline
        \end{tabular}
    }
    \caption{\textbf{Ablation on architecture of Unified Token Transformer (\textbf{\textit{UTT}}).} \textbf{Ours (gru)} means the architecture of \textbf{\textit{UTT}} is GRU-based, and \textbf{Ours (gpt)} means we use GPT-based architecture for \textbf{\textit{UTT}}. Both architectures adopt a deterministic token prediction  strategy, indicating no injection of $\textit{z}\thicksim\mathcal{N}(0, I)$ during inference.}
    \label{tab:ablation-gpt-gru}
\end{table}

\begin{table*}[ht]
    \centering
    \resizebox{0.85\linewidth}{!}{
        \begin{tabular}{ccccccc|cccc}
            \hline
            \multirow{2}{*}{Method} & \multicolumn{6}{c|}{Text-to-Motion} & \multicolumn{4}{c}{Audio-to-Motion} \\
            \cline{2-11}
             & Top-1 Acc. $\uparrow$ & Top-5 Acc. $\uparrow$ & $\rm FID_K$ $\downarrow$ & $\rm FID_m$ $\downarrow$ & $\rm Div_k$ $\uparrow$ & $\rm Div_m$ $\uparrow$ & $\rm FID_k$ $\downarrow$ & $\rm FID_m$ $\downarrow$ & $\rm Div_k$ $\uparrow$ & $\rm Div_m$ $\uparrow$ \\
            \hline
            ours & 8.11 & 25.01 & 27.66 & 4.92 & 4.28 & 6.77 & 28.44 & 15.70 & 6.13 & 4.07 \\
            \rowcolor[rgb]{0.902,0.898,0.898} ours+\textit{z} & \textbf{8.33} & \textbf{25.78} & \textbf{26.93} & \textbf{4.49} & \textbf{5.05} & \textbf{7.63} & \textbf{27.99} & \textbf{15.26} & \textbf{6.32} & \textbf{4.14} \\
            \hline
        \end{tabular}
    }
    \caption{\textbf{Ablation on diversity at token prediction.} We validate the performance our Unified Token Transformer (\textbf{\textit{UTT}}) with or without injecting $\textit{z}\thicksim\mathcal{N}(0, I)$, where \textbf{ours} means deterministic prediction, and \textbf{ours+\textit{z}} means probabilistic prediction mode. For \textbf{ours+\textit{z}}, we generate 30 samples per input and measure the average metrics.}
    \label{tab:ablation-gpt-gpt(z)}
\end{table*}

\begin{table*}[ht]
    \centering
    \resizebox{0.85\linewidth}{!}{
        \begin{tabular}{ccccccc|cccc}
            \hline
            \multirow{2}{*}{Method} & \multicolumn{6}{c|}{Text-to-Motion} & \multicolumn{4}{c}{Audio-to-Motion} \\
            \cline{2-11}
             & Top-1 Acc. $\uparrow$ & Top-5 Acc. $\uparrow$ & $\rm FID_k$ $\downarrow$ & $\rm FID_m$ $\downarrow$ & $\rm Div_k$ $\uparrow$ & $\rm Div_m$ $\uparrow$ & $\rm FID_k$ $\downarrow$ & $\rm FID_m$ $\downarrow$ & $\rm Div_k$ $\uparrow$ & $\rm Div_m$ $\uparrow$ \\
            \hline
            vq-decoder & 8.11 & 25.01 & 27.66 & 4.92 & 4.28 & 6.77 & 28.44 & 15.70 & 6.13 & 4.07 \\
            \rowcolor[rgb]{0.902,0.898,0.898} diffusion-decoder & \textbf{8.12} & \textbf{25.18} & \textbf{23.93} & \textbf{3.15} & \textbf{6.98} & \textbf{7.17} & \textbf{18.43} & \textbf{10.39} & \textbf{6.84} & \textbf{5.03} \\
            \hline
        \end{tabular}
    }
    \caption{\textbf{Ablation on diversity at token decoding.} We compare the performance of token decoding using our Diffusion Motion Decoder (\textbf{\textit{DMD}}) and pretrained VQ-Decoder. For fair comparison, both experiments take input as the token sequence predicted by our GPT-based \textbf{\textit{UTT}} without injecting $\textit{z}\thicksim\mathcal{N}(0, I)$. For $\textbf{\textit{DMD}}$, we generate 30 samples per input.}
    \label{tab:ablation-vqdecoder-diffusion}
\end{table*}

\subsection{Datasets}
As there is no public dataset that supports text-driven and audio-driven motion generation simultaneously, we use two separate datasets in our experiment. The first dataset is HumanML3D\cite{Guo_2022_CVPR}, which is a text-to-motion dataset built upon AMASS dataset\cite{mahmood2019amass} and HumanAct12\cite{guo2020action2motion}. It provides a wide range of motion-language pairs which cover ordinary activities, such as `jumping', `walking', `running', etc. The second dataset is AIST++\cite{li2021ai}, a large-scale dance motion dataset built from\cite{aist-dance-db}. It contains 1409 sequences of dance motions, covering 10 different dance genres with hundreds of choreographies.\par

\subsection{Implementation Details}

\paragraph{Data Preprocessing.} The raw motion format of HumanML3D follows the SMPL skeleton with 22 joints, while the format of AIST++ follows the SMPL skeleton with 24 joints, preprocessing is conducted to unify their format. For each dataset, we use \cite{SMPL-X:2019} to convert their motion representation to SMPL skeleton with 24 joints representation. Furthermore, we normalize each motion sequence by transforming the initial pose heading toward the same direction. For audio preprocessing, a public toolbox, Librosa \cite{jin2017towards}, is used to extract the audio features. The feature consists of Mel Frequency Cepstral Coefficients (MFCC), MFCC delta, constant-Q chromagram, tempogram, and onset strength. For each audio feature sequence, it is represented as a $\textit{T} \times 438$ matrix.
\vspace{-4mm}

\paragraph{Motion Quantization.} The codebook size is set to $2048 \times 1024$, where the number of discrete tokens is 2048, and the dimension of each token is 1024. For VQ-encoder and VQ-decoder, three-layer temporal 1D convolution networks are adopted. We set $\beta_1 = 1$, and $\beta_2 = 1$.
\vspace{-4mm}
\paragraph{Modality-Agnostic Transformer Encoder.}For the text encoder, we use the pre-trained CLIP text encoder, for audio encoder, a 1-layer FC is adopted. For inputs of both modalities, we project them to the dimension of 256. The number of transformer encoder layers is set to 6, the number of attention heads is 8 and the hidden dimension is 1024.
\vspace{-4mm}

\paragraph{Unified Token Transformer.}We set the number of transformer encoder layers to 8 and set the hidden dimension to 1024. For the loss, we set $\beta_{adv}=1$.
\vspace{-4mm}

\paragraph{Diffusion Motion Decoder.} For condition encoder $\mathcal{E}_{z}$, the number of  encoder layers is 8, and the number of layers of the decoder is also set to 8. For both encoder and decoder, the hidden dimension is 1024, and the number of attention heads is 8. The number of diffusion steps is set to 1000 in our experiments.
\vspace{-4mm}

\paragraph{Learning rate and Optimizer.}For all stages, we use Adam as our optimizer with a learning rate of 0.0001.

\subsection{Evaluation Metrics}

\paragraph{Text-to-Motion Evaluation Metrics.}
We evaluate our method on text-to-motion tasks with four types of metrics. \textbf{1}) $\textit{Text Retrieval Accuracy}$. We evaluate the correlation between motion sequence and text description by Top-1 Acc, and Top-5 Acc. respectively. Following \cite{tevet2022motionclip}, we train a motion encoder with a contrastive learning paradigm to make the embedding of paired motion and text description close to each other. During the evaluation, we compute the embedding of paired motion and text description, and another randomly select 60 irrelevant text descriptions from testset. Then we calculate the similarity between motion embedding and 61 text embedding. If the paired text's embedding is the most similar one among these text embedding, it is considered as Top-1 Acc., similar for Top-5 Acc. \textbf{2}) $\textit{Frechet Inception Distance (FID)}$. FID measures the similarity between two distributions, and we use a pre-defined model to extract features from ground truth samples and generated samples, and measure the similarity between them. Here we define ${\rm FID}_k$ to measure the kinetic feature, and ${\rm FID}_m$ to measure the manual defined geometric feature. \textbf{3}) $\textit{Diversity}$. We measure the diversity of the kinetic and geometric features. The features are extracted by the same model as FID does. \textbf{4}) $\textit{Reconstruction Accuracy}$. We measure the average joints positions distance and root trajectory distance\cite{petrovich2022temos} between ground truth samples and generated samples to indicate how close the generated samples are to the ground truth samples in terms of pose geometry.\par

\begin{figure}[ht]
    \centering
    \includegraphics[width=\linewidth]{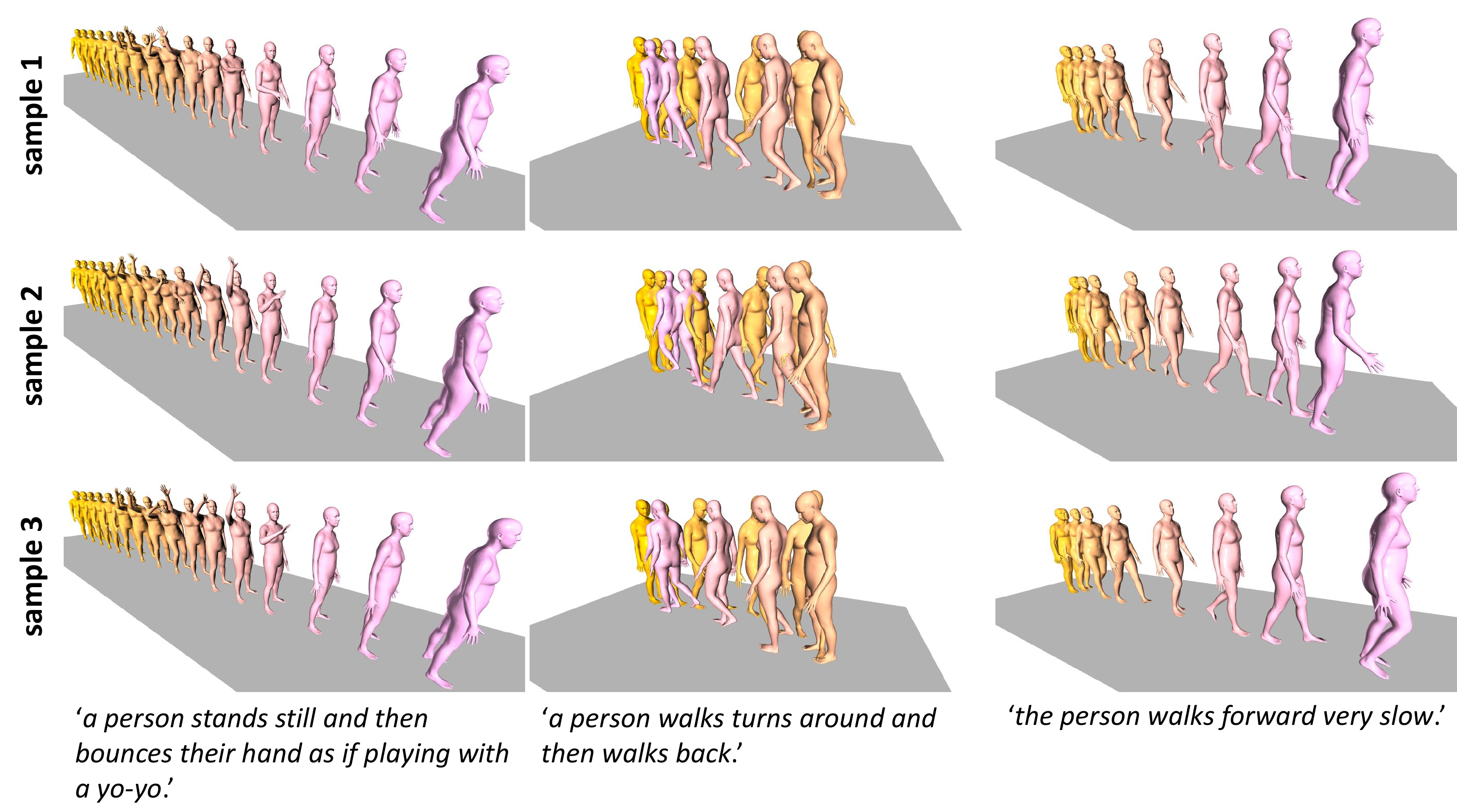}
    \caption{\textbf{Diversity of Text-to-Motion.} Each column shows 3 samples generated by the same text descriptions. We observe that our method generates motion sequences with high diversity while maintaining the semantic meaning of the driving text descriptions.}
    \label{fig:t2m-diversity}
\end{figure}

\begin{figure}[ht]
    \centering
    \includegraphics[width=\linewidth]{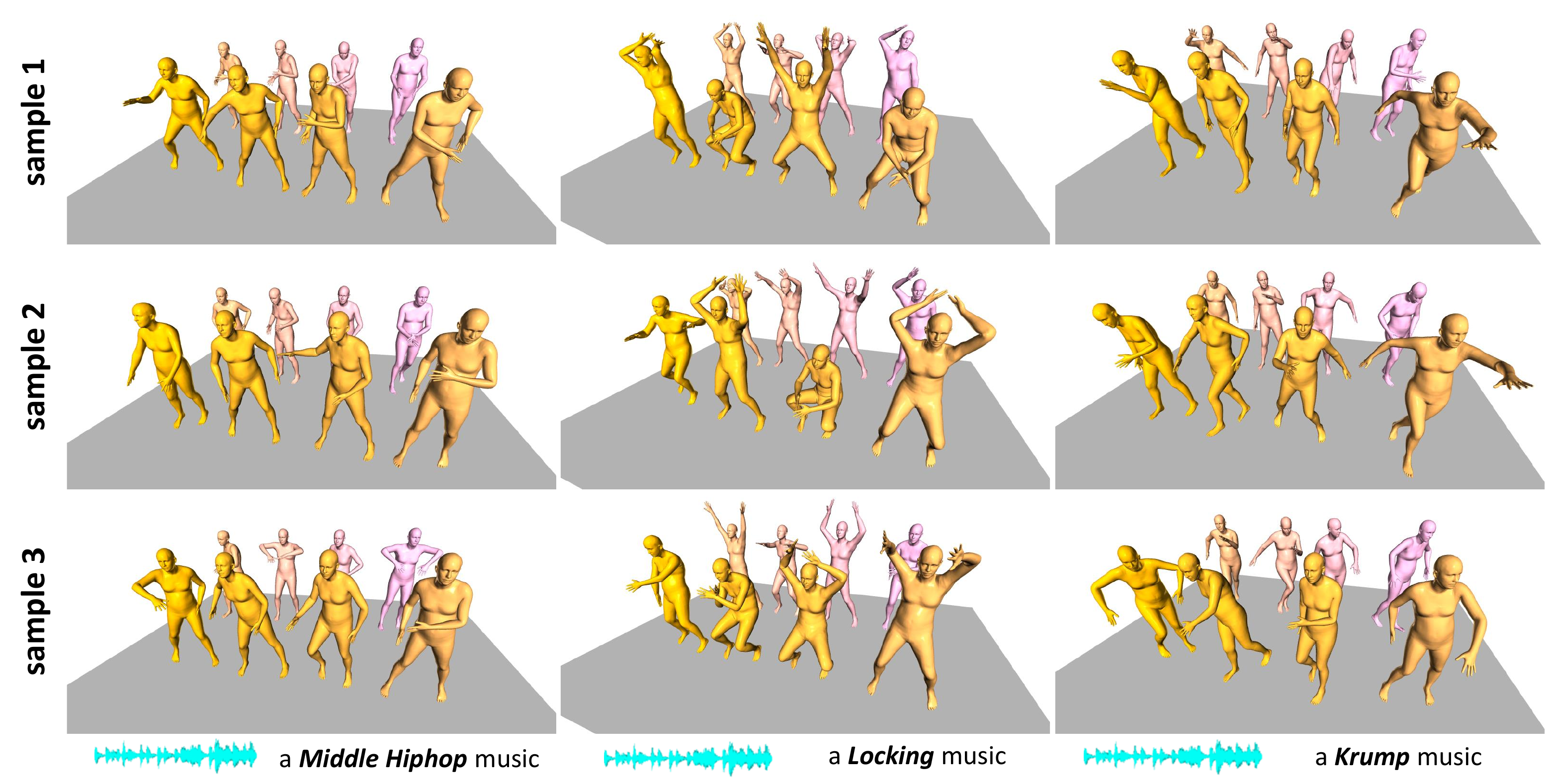}
    \caption{\textbf{Diversity of Audio-to-Motion.} We show results driven by 3 different music clips. 3 samples are generated for each clip. For better visualization, we evenly extract 8 frames per sample and display them in a grid format. As observed, our method generates audio-driven motion sequences with high diversity.}
    \label{fig:a2m-diversity}
\end{figure}
\vspace{-3mm}
\paragraph{Audio-to-Motion Evaluation Metrics.}
We evaluate our method on audio-to-motion tasks with four types of metrics. \textbf{1}) $\textit{Beat Align}$. We measure how close are generated motion is to the driving audio sequence in terms of rhythmic beat. The Beat Alignment Score is calculated as: $\frac{1}{|\textit{B}^a|}\sum_{\textit{t}^a \in \textit{B}^a} \exp{\left\{-\frac{\min_{t^m \in B^m}\|t^m - t^a\|^2}{2\sigma^2}\right\}}$,
\noindent where $\textit{B}^m$ and $\textit{B}^a$ correspond to the motion beat and audio beat, respectively. For $\textit{FID}$, \textit{Diversity}, and \textit{Reconstruction Accuracy}, we follow the same definition as described in the text-to-motion section.\par


\subsection{Results}
We compare our method with several state-of-the-art methods. For text-to-motion, we compare ours with TEMOS\cite{petrovich2022temos}, MotionCLIP\cite{tevet2022motionclip}, and TM2T\cite{guo2022tm2t}. For audio-to-motion, we compare with FACT\cite{li2021ai} and Bailando\cite{siyao2022bailando}. 
\paragraph{Quantitative Comparison.}Tab. \ref{tab:text-to-motion-table} and Tab. \ref{tab:audio-to-motion-table} summarize the quantitative comparison results. $\textbf{1}$) For the text-to-motion task, Tab. \ref{tab:text-to-motion-table} shows that our method outperforms all the competitive prior methods on $\textit{Text Retrieval Acc.}$ and $\textit{Recon Acc.}$ metrics. Specifically, our results significantly outperform all prior methods on $\textit{Text Retrieval Acc.}$, which indicates our method generates semantically more correlated samples among all of them. Thanks to our probabilistic token prediction($\textbf{UTT}$) and token decoding($\textbf{DMD}$), our method achieves the significantly higher kinetic feature diversity among all methods. $\textbf{2}$) Tab. \ref{tab:audio-to-motion-table} shows that our method achieves much better results in terms of ${\rm FID}_k$ and ${\rm FID}_m$. The significant improvement of $\textit{FID}$ scores of ours indicating the samples generated by our method show best kinetic and geometric quality among all. Also, ours achieves second best on ${\rm Div}_k$ score, slightly lower than \cite{siyao2022bailando}. It indicates that our method is also able to generate audio-driven motion with competitive kinetic diversity.
\vspace{-4mm}
\paragraph{Qualitative Comparison.}Fig.~\ref{fig:compare} shows the qualitative comparison between ours and the prior methods. $\textbf{1}$) For the text-to-motion task, our method shows the best quality and semantic correlation. \ie, for the sample-driven by text \textit{'a person raises their arms out to their sides lowers them back down and then leading with their left arm and leg gets down on their hands and knees before raising up and standing on their knees.'}, which describes a very complex scenario, the results generated by \cite{petrovich2022temos} fails to show the action $\textit{'standing on their knees'}$, and \cite{tevet2022motionclip} only shows the action $\textit{'raises their arms out to their sides'}$, the results of \cite{guo2022tm2t}, similarly, only shows the action $\textit{'raises their arms out to their sides'}$, while ours shows the full action sequence described in the text description.  $\textbf{2}$) For the audio-to-motion task, ours also shows the highest quality and correlation compared with \cite{li2021ai} and \cite{siyao2022bailando}. For example, when driven by $\textbf{\textit{Locking}}$ music, the motion sequence of \cite{siyao2022bailando} repeats a similar pattern, and the motion sequence of \cite{li2021ai} looks like the person is dancing to a Ballet style music, but not a Locking style music, while ours shows diverse poses and maintains the high correlation to the driving music style at the same time.\par
\vspace{-4mm}
\paragraph{Diversity.}Fig. \ref{fig:t2m-diversity} and Fig. \ref{fig:a2m-diversity} show the qualitative results of our model on text-to-motion and audio-to-motion task, respectively. For each figure, we show 3 samples generated by the same driving input. We can observe from the visualization results that our method generates motion sequences with high diversity while maintaining semantic correlation as well. For example the samples driven by text \textit{'the person walks forward very slowly'}. The motion sequences correctly match the text input while displaying different details even for such simple scenario. Fig. \ref{fig:a2m-diversity} shows the evenly cropped poses of the  results of the audio-driven task. We show results driven by 3 different types of music, and for each music sequence, 3 samples are shown. As we can see, our method generates motion sequences with high quality and diversity.\par

\subsection{Ablation Study}
\setlength{\parindent}{1em}We conduct ablation studies on ($\textbf{1}$) design of $\textbf{UTT}$, ($\textbf{2}$) deterministic v.s. probabilistic generation of $\textbf{UTT}$, and ($\textbf{3}$) diversity of $\textbf{DMD}$ compared with VQ-Decoder.\par
\vspace{-4mm}

\paragraph{Variants of Unified Token Transformer.}We explore the design of $\textbf{UTT}$ with two different architectures, GRU-based and GPT-based, where GPT-based is the model adopted in our work. We don't inject random noise $\textit{z}\thicksim\mathcal{N}(0, I)$ to eliminate randomness. Tab. \ref{tab:ablation-gpt-gru} shows the quantitative comparison between these designs. We can observe that the GPT-based achieves better results on both text-to-motion task and audio-to-motion task. For text-to-motion task, GPT-based method achieves higher text retrieval accuracy, meaning that the samples are more semantically correlated to the input text. In addition, GPT-based method also outperforms on $\textit{FID}$s remarkably on audio-to-motion task, indicating that its capacity in generating more realistic motion to audio. This is likely because a token can attend to all its previous tokens in GPT-based approach, making its long-term generation more stable and semantically correlated.\par
\vspace{-4mm}

\paragraph{Diversity at Unified Token Transformer.}We explore the diversity of token prediction of Unified Token Transformer $\textbf{UTT}$. We adopt the VQ-Decoder as our token decoder to eliminate the influence of diversity at token decoding to our final results. For probabilistic prediction mode($\textbf{ours+\textit{z}}$), we generate 30 samples for each input and measure the average metrics. And for deterministic prediction mode($\textbf{ours}$), we only generate 1 sample for each input. Tab. \ref{tab:ablation-gpt-gpt(z)} summarizes the quantitative results. It shows that our $\textbf{UTT}$ with probabilistic generation achieves better results compared with deterministic generation. Specifically, on text-to-motion task, $\textbf{ours+\textit{z}}$ outperforms on Top-1 Acc. and Top-5 Acc., and also achieves higher ${\rm Div}_k$ and ${\rm Div}_m$ scores, respectively, as expected. For audio-to-motion task, $\textbf{ours+\textit{z}}$ also achieves better results compared with $\textbf{ours}$. For ${\rm FID}_k$ and ${\rm FID}_m$ metrics, probabilistic generation mode brings better kinetic quality (28.44 vs 27.99) and geometric quality (15.70 vs 15.26). It also shows that injecting $\textit{z}\thicksim\mathcal{N}(0,I)$ to token prediction stage brings slightly higher kinetic diversity ${\rm DIV}_k$ and geometric diversity ${\rm DIV}_m$, respectively. The quantitative comparison shows that injecting $\textit{z}\thicksim\mathcal{N}(0, I)$ to $\textbf{UTT}$ not only brings motion quality gain but also brings higher diversity, as expected.\par
\vspace{-4mm}

\paragraph{Diversity at Diffusion Motion Decoder.}We explore the performance of our $\textbf{DMD}$ at token decoding in terms of the motion quality and diversity, compared with VQ-Decoder. In this experiment, we adopt the GPT-based $\textbf{UTT}$ and don't inject $\textit{z}\thicksim\mathcal{N}(0, I)$ to eliminate randomness at token prediction stage. We compare the motion decoding between our probabilistic generation module $\textbf{DMD}$ and deterministic generation module VQ-Decoder. Tab. \ref{tab:ablation-vqdecoder-diffusion} summarizes the comparison results. For the text-to-motion task, we can observe that $\textbf{DMD}$ achieves slightly higher text retrieval accuracy. It also outperforms on ${\rm FID}_k$, ${\rm FID}_m$, ${\rm Div}_k$ and ${\rm Div}_m$ largely. The similar trends is observed on audio-to-motion task, where $\textbf{DMD}$ outperforms on ${\rm FID}_k$ and ${\rm FID}_m$ significantly. For ${\rm FID}_k$, $\textbf{DMD}$ reduces it by 35\%(28.44 to 18.43), and for ${\rm FID}_m$, $\textbf{DMD}$ also brings 33\% quality gain(15.70 to 10.39). In terms of $\textit{Diversity}$, $\textbf{DMD}$ also brings noticeable performance gain. The comparison suggests that our $\textbf{DMD}$ design plays a vital role in generating diversity while maintaining high sample quality. It is especially obvious that this design performs better on audio-to-motion task.\par
\vspace{-4mm}

\section{Discussion}
\setlength{\parindent}{1em}In this paper, we propose a Unified Driving Engine for human motion generation. Our method unifies text-driven and audio-driven human motion generation tasks into one model. We learn a semantic-rich codebook to represent various patterns of human motions and propose a Modality-Agnostic Transformer Encoder to map inputs of different modalities into a joint space. We propose a Unified Token Transformer to predict motion tokens with high diversity, and finally, we propose Diffusion Motion Decoder to bring additional diversity to the token decoding process. Experiments show that our method achieves state-of-the-art performance on text-to-motion and audio-to-motion tasks, respectively. Our current method can be regarded as a late fusion mechanism, it would be interesting to explore an early fusion between different modalities in future work.\par
\vspace{-4mm}


{\small
\bibliographystyle{ieee_fullname}
\bibliography{egbib}
}

\clearpage
\appendix

\twocolumn[
\begin{@twocolumnfalse}
\section*{\centering{Supplementary Material}}
\end{@twocolumnfalse}
]

\section{Unified \vs Modality-Specific}
We first provide more justification for unifying training of two tasks, text-to-motion, and audio-to-motion, as opposed to training modality-specific models.\par

\begin{figure}[ht]
    \centering
    \includegraphics[width=0.8\linewidth]{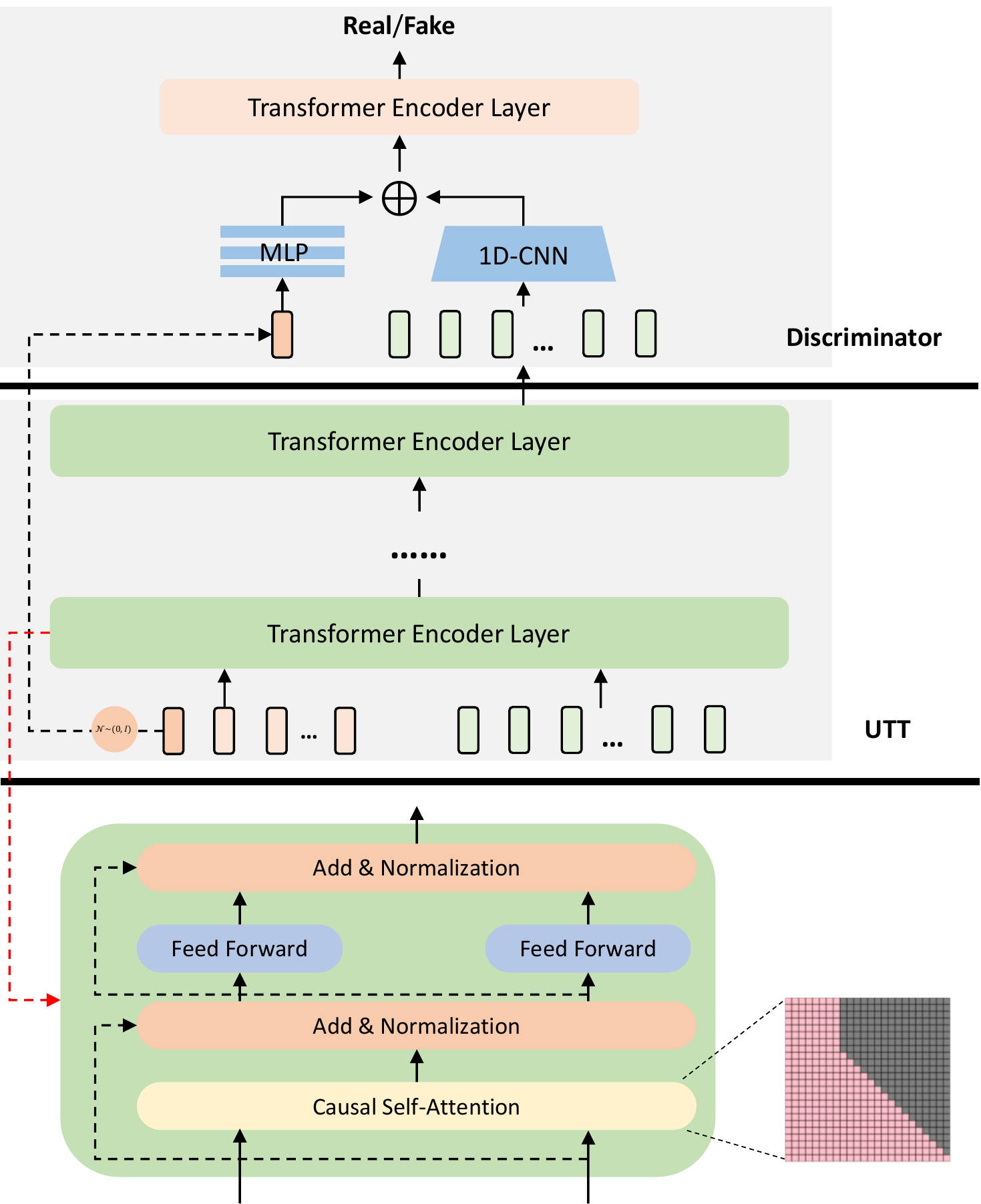}
    \caption{\textbf{Architecture of UTT.} \textbf{Bottom} panel shows detail of the transformer encoder layer in \textbf{middle} panel, where a causal self-attention is adopted in replacing with conventional full self-attention. The \textbf{top} panel is the detail of the conditional discriminator. The global embedding and predicted token sequences are transformed and summed together and fed to a transformer encoder to get a patch-wise validity score.}
    \label{fig:supp_utt}
\end{figure}

\begin{figure}[ht]
    \centering
    \includegraphics[width=0.8\linewidth]{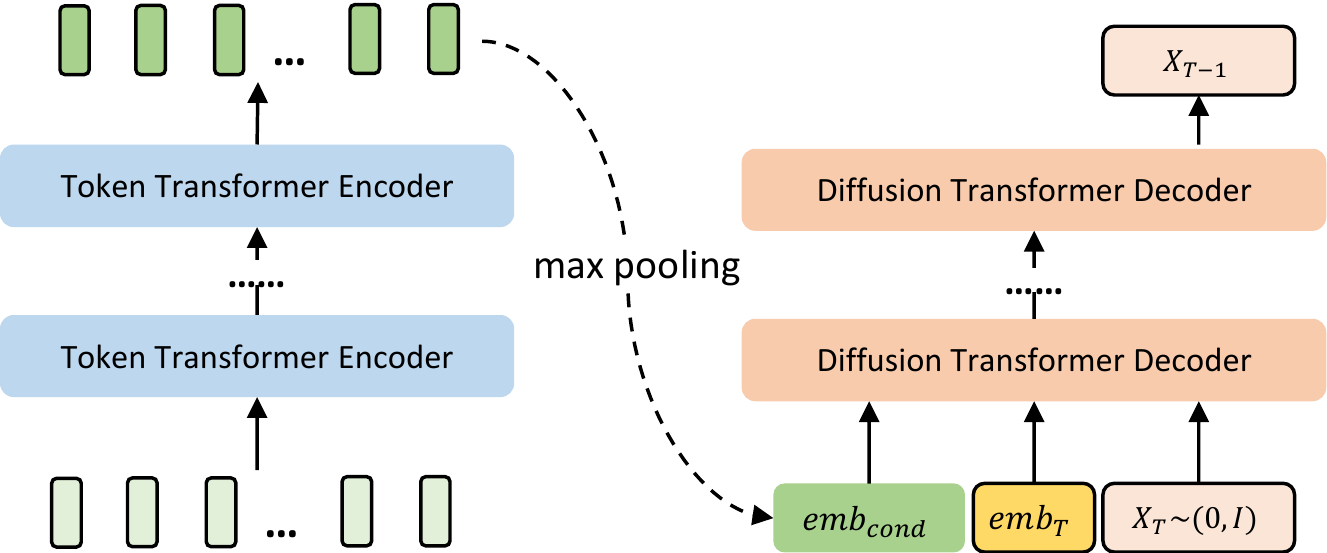}
    \caption{\textbf{Architecture of DMD.} \textbf{Left} panel shows the detail of Token Transformer Encoder. The predicted token sequences are fed to this module, and the sequential embedding is obtained. Then we get a single embedding by max-pooling operation along the temporal dimension. This single embedding is used as a condition at the Diffusion Transformer Decoder module. \textbf{Right} panel shows the detail of the Diffusion Transformer Decoder. We feed the condition embedding, timestep embedding, and latent to it, and the gaussian noise $\epsilon_t$ is obtained at each reversed diffusion step.}
    \label{fig:supp_dmd}
\end{figure}

\subsection{Unification Brings Smooth Transition of Generated Motion between Modalities}
\setlength{\parindent}{1em}We justify why we choose to unify the two tasks in one shared model by showing qualitative examples of driving motion sequences with both text descriptions and audio sequences as input and providing a brief analysis based on the experimental results. To evaluate, we first feed text descriptions to our model to generate a motion sequence, then feed both an audio clip and the last 8 tokens of the generated motion sequence as primitive to our model to generate subsequent motion sequences. By feeding both the audio sequence and the last tokens of the motion sequence, we are hoping that the generated motion will be conditioned on both audio input and motion primitives, simultaneously, and then a smooth transition can be obtained from one to another naturally.\par

\begin{figure*}[ht]
    \centering
    \includegraphics[width=\textwidth]{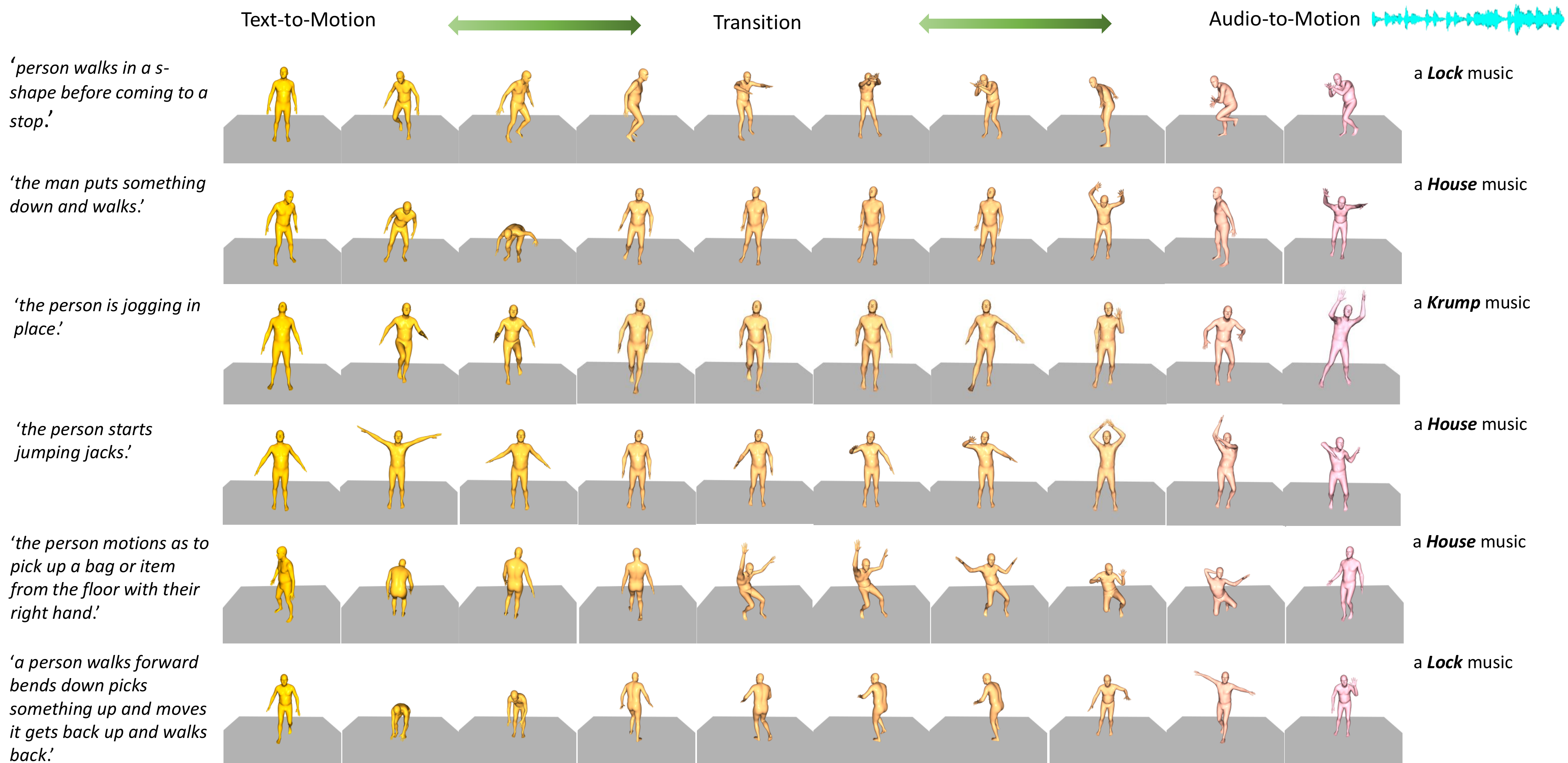}
    \caption{\textbf{Examples of unified driven samples.}Each row shows a motion sequence driven by a text description and an audio clip sequentially. The text descriptions are fed to UDE first to generate a text-conditioned motion sequence. Then the audio clip and the last 8 tokens of text-conditioned motion are fed to UDE to generate an audio-conditioned motion sequence. For each row, we extracted 10 frames sequentially to show the transition between text-driven and audio-driven sequences. Demo videos could be found in our supplementary materials.}
    \label{fig:supp_unified}
\end{figure*}

\paragraph{Qualitative Examples}\label{sec:supp_para_1} Fig. \ref{fig:supp_unified} shows results of much more complex motion sequences driven by text descriptions and audio clips sequentially. There are 3 regions shown in each row of the figure. For $\textit{Text-to-Motion}$ region, the motion is controlled by text description dominantly, while for $\textit{Audio-to-Motion}$ region, the motion is driven by audio clip mainly. The $\textit{Transition}$ region, in the middle of each row, shows how the text-driven motion sequence smoothly transits to an audio-driven sequence without introducing additional motion in-between modules.\par

\setlength{\parindent}{1em}The qualitative results suggest that complex scenarios correlated with multimodal inputs could be generated by our UDE model without introducing additional motion in-between modules. We show that by feeding the multimodal input to our model sequentially, and by conditioning the current task on previously predicted tokens, we can generate more complex motion sequences smoothly transit from one scenario(text) to another(audio).\par

\setlength{\parindent}{1em}Please refer to our supplementary video for better visualization of the results of such mixed input driving tasks.\par

\paragraph{Analysis} Sec.\ref{sec:supp_para_1} shows why we propose to unify these tasks visually, we give a brief analysis of this problem here. As stated in sec. \ref{sec:supp_para_1}, we feed text to generate motion sequence at first, which we denote as $\tilde{x}^t=\mathcal{D}_{DMD}(\mathcal{E}_{UTT}(e^t))$, here $\mathcal{D}_{DMD}(\cdot)$ is the Diffusion Motion Decoder $\tilde{x}^t=\mathcal{D}_{DMD}(z^q)$, $z^q$ is the motion token sequence, and $\mathcal{E}_{UTT}(\cdot)$ is the Unified Token Transformer which maps embedding to motion token sequence as $\textit{z}^q=\mathcal{E}_{UTT}(e^t)$. Let's denote the last $n$ tokens of $z^q$ as $z^{q, T-n:T}$, where $T$ is the length of the token sequence. Then we feed an audio clip and the last $n$ tokens $\textit{z}^{q,T-n:T}$ to generate motion sequence as $\tilde{x}^a=\mathcal{D}_{DMD}(\mathcal{E}_{UTT}(e^a, z^{q, T-n:T}))$. In this step, we notice that the input to UTT $\mathcal{E}_{UTT}(\cdot)$ has two items, 1) the first item is the embedding of audio clip $e^a$, and 2) the second term is the last $n$ tokens $z^{q, T-n:T}$ which corresponds to text description. If we adopt a modality-specific paradigm, $\mathcal{E}_{UTT}(e^a, z^{q, T-n:T})$ will give unexpected results because $\mathcal{E}_{UTT}(\cdot)$ is trained either on text modality only or audio modality only. However, in our setting, the input to $\mathcal{E}_{UTT}(\cdot)$ covers two modalities, the embedding $e^a$ corresponds to audio modality, and $z^{q, T-n:T}$ corresponds to text modality because it is obtained by text description, and vise versa. To conclude, the codebook corresponding to different scenario will not be shared in Motion Quantization and Unified Token Transformer modules, hindering the token prediction conditioned on cross modality scenario. As a consequence, a model trained on a modality-specific paradigm will not perform well in generating smoothly transited motion driven by one modality to another.

\subsection{Unification Brings Strong Results \& Engineering Efficiency}
\setlength{\parindent}{1em} Here we provide more quantitative analysis. To compare, we also train  modality-specific models on text-driven and audio-driven tasks, respectively, and separately. \cmmnt{For text-to-audio, we still train our model on \cmmnt{\cite{Guo_2022_CVPR}}, and for audio-to-motion task, we train our model on \cmmnt{\cite{li2021ai}}.} We keep the model architecture fixed for both Modality-Agnostic Transformer Encoder(\textbf{MATE}), and Unified Token Transformer(\textbf{UTT}), and we don't train them with Diffusion Motion Decoder(\textbf{DMD}) because we don't want to introduce diversity at this time for a fair comparison. Therefore, we just adopt the pretrained VQ-Decoder in Motion Quantization(\textbf{MQ}) stage. For both text-to-motion and audio-to-motion tasks, we follow the same optimization strategy described above and trained 300 epochs for each task. During the evaluation, we follow the deterministic token prediction strategy described above, where we don't inject $\textit{z}\thicksim\mathcal{N}(0, I)$ to $\textbf{UTT}$ because diversity is not desired at this stage. We evaluate the performance of our $\textbf{UDE}$ model over the same metrics as above: 1) For text-to-motion, we evaluate our method on $\textit{Text Retrieval Acc.}$, $\textit{FID}$ scores, and $\textit{Diversity}$. 2) For audio-to-motion, we evaluate our method on $\textit{Beat Align Score}$, $\textit{FID}$, and $\textit{Diversity}$, respectively.\par

\setlength{\parindent}{1em}Tab. \ref{tab:supp_tab} summarizes the quantitative results. As we can observe from the results, for the text-to-motion task, training our model on text-to-motion dataset\cmmnt{\cite{Guo_2022_CVPR}} only does not bring obvious performance gain. On the contrary, training a text-only model brings even worse Top-1 Acc. and $\textbf{FID}$s noticeably. For Top-1 Acc. text-only training brings around 5\% accuracy drop against unified training (8.81 to 7.77). If we take a look at the $\textbf{FID}$s, text-only training also brings worse results. A similar conclusion could be drawn from the audio-to-motion task. If we train an audio-only model, it does not improve performance. For $\textit{Beat Align}$, unified training shows a slightly better beat synchronization property. For feature-wise quality and diversity, audio-only and unified models draw a tie. For audio only model, it has better performance on ${\rm FID}_m$ and ${\rm Div}_m$, while for the unified model, better kinetic quality ${\rm FID}_k$ and kinetic diversity ${\rm Div}_k$ are obtained.\par

\setlength{\parindent}{1em}This study suggests that our model achieves competitive performance on both text-driven and audio-driven scenarios when trained in a unified paradigm, compared with training on the uni-task paradigm. With the performance maintained, our method successfully puts these two driven tasks to one unified solution. Through unification, engineering efficiency is improved because only one model needs to be maintained and improved for possible future applications. This suggests there is potential for unification on multimodal human motion generation.\par

\begin{table*}[ht]
    \centering
    \resizebox{\linewidth}{!}{
        \begin{tabular}{ccccccc|ccccc}
            \hline
            \multirow{3}{*}{Method} & \multicolumn{6}{c}{Text-to-Motion} & \multicolumn{5}{|c}{Audio-to-Motion} \\
            \cline{2-12}
            & \multicolumn{2}{c}{Text Retrieval Acc.} & \multicolumn{2}{c}{FID} & \multicolumn{2}{c|}{Diversity} & \multirow{2}{*}{Beat Align $\uparrow$} & \multicolumn{2}{c}{FID} & \multicolumn{2}{c}{Diversity} \\
            & Top-1 Acc. $\uparrow$ & Top-5 Acc. $\uparrow$ & $\rm FID_k$ $\downarrow$ & $\rm FID_m$ $\downarrow$ & $\rm Div_k$ $\uparrow$ & $\rm Div_m$ $\uparrow$ & & $\rm FID_k$ $\downarrow$ & $\rm FID_m$ $\downarrow$ & $\rm Div_k$ $\uparrow$ & $\rm Div_m$ $\uparrow$ \\
            \hline
            text only & 7.77 & \textbf{26.01} & 31.73 & 5.69 & \textbf{4.42} & \textbf{6.96} & - & - & - & - & - \\
            audio only & - & - & - & - & - & - & 0.2231 & 39.27 & \textbf{11.65} & 5.68 & \textbf{8.03} \\
            unified & \textbf{8.11} & 25.01 & \textbf{27.66} & \textbf{4.92} & 4.28 & 6.77 & \textbf{0.2268} & \textbf{28.44} & 15.70 & \textbf{6.13} & 4.07 \\
            \hline
        \end{tabular}
    }
    \caption{\textbf{Ablation on Unification v.s. Modality-Specific.} We explore our method trained in a unified paradigm against that trained in a modality-specific paradigm. All three models, use exactly the same architecture, and we don't inject any random term to eliminate the influence of diversity. For $\textbf{text only}$ and $\textbf{audio only}$, they are trained on HumanML3D and AIST++ datasets solely.}
    \label{tab:supp_tab}
\end{table*}

\begin{figure*}[ht]
    \centering
    \includegraphics[width=\textwidth]{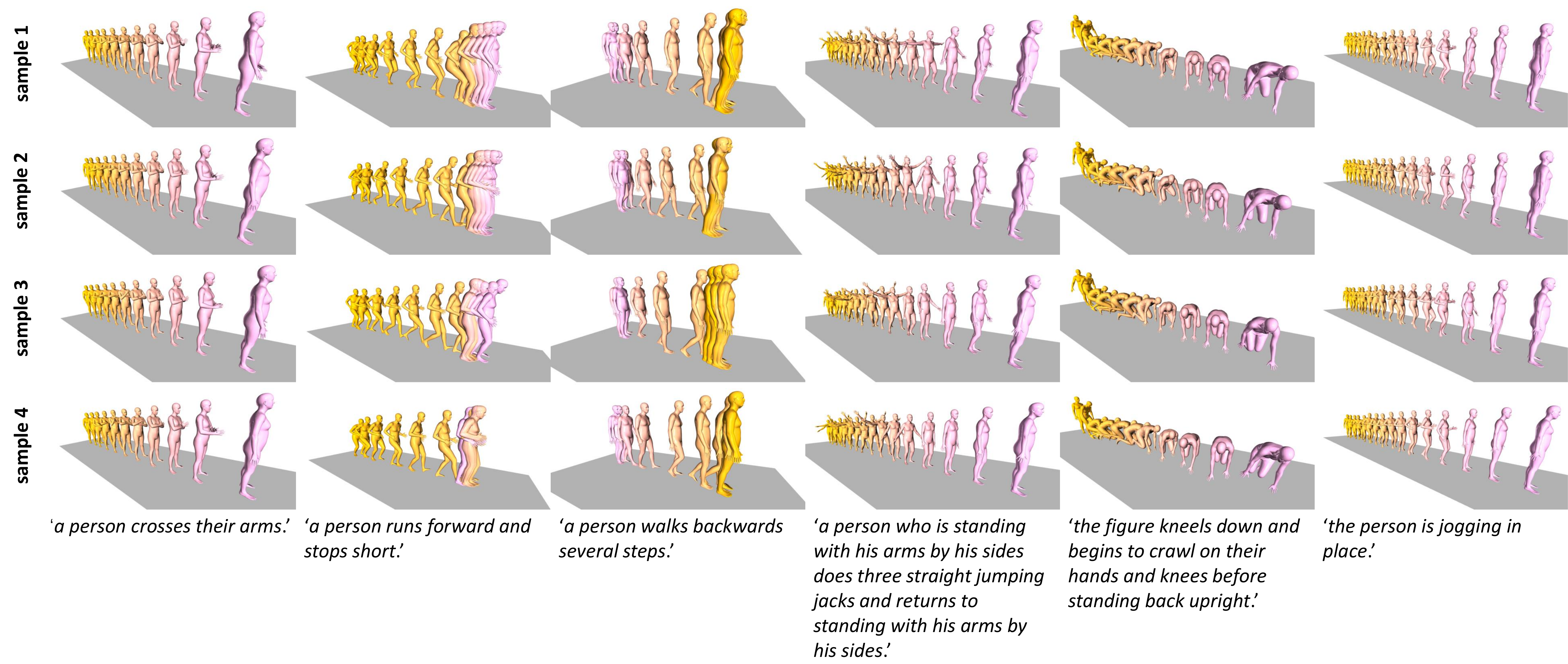}
    \caption{\textbf{Diversity of Text-to-Motion.} We show more qualitative results on text-to-motion tasks. For each column, we show 4 samples driven by the same text description with high diversity. We adjust the trajectory of some motion sequences for better visualization. Demo videos could be found in our supplementary materials.}
    \label{fig:supp_t2m_diversity}
\end{figure*}

\begin{figure*}[ht]
    \centering
    \includegraphics[width=\linewidth]{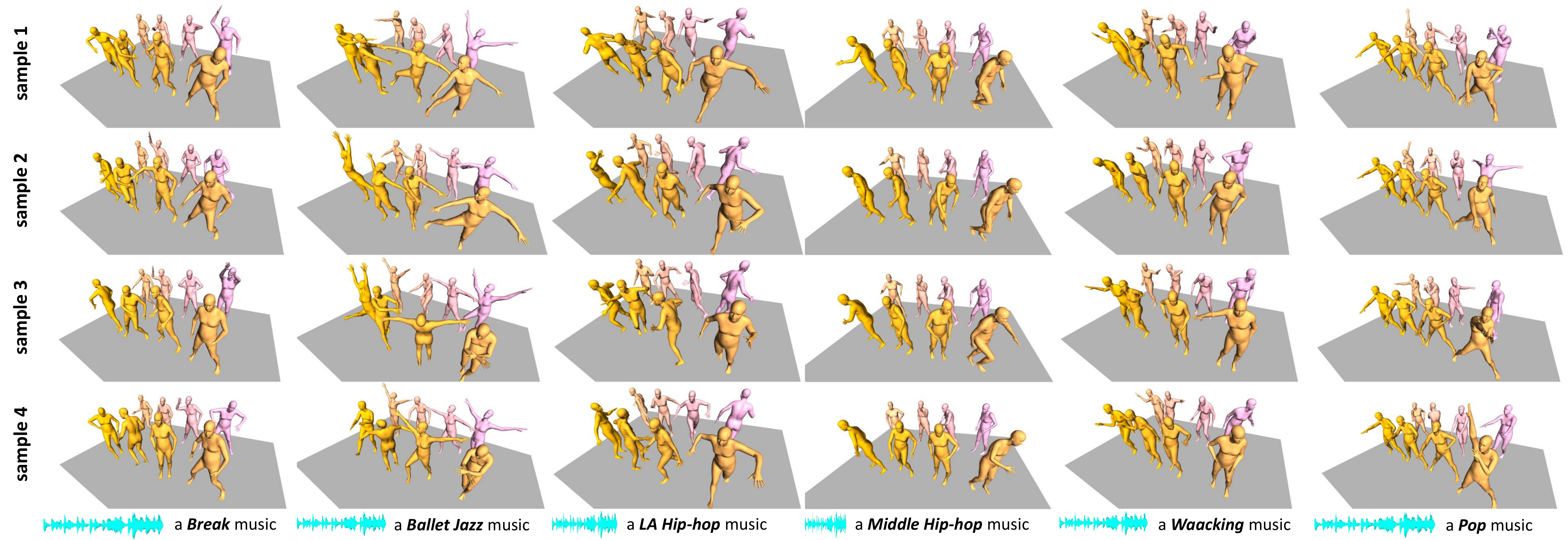}
    \caption{\textbf{Diversity of Audio-to-Motion.} We show more qualitative results on audio-to-motion tasks. For each column, we show 4 samples driven by the same audio clip with high diversity. We show samples driven by 6 audio clips with different genres. Demo videos could be found in our supplementary materials.}
    \label{fig:supp_a2m_diversity}
\end{figure*}

\section{Detail of Model Architecture}
\setlength{\parindent}{1em}We describe the detail architecture of Unified Transformer Encoder($\textbf{UTT}$) and Diffusion Motion Decoder($\textbf{DMD}$) here. Fig. \ref{fig:supp_utt} describes the architecture of $\textbf{UTT}$. We describe the Unified Token Transformer module and the conditional discriminator in an end-to-end manner, and the transformer encoder layer with causal self-attention is demonstrated at the bottom panel of Fig. \ref{fig:supp_utt}. The detailed architecture of $\textbf{DMD}$ is shown in Fig. \ref{fig:supp_dmd}, where the left panel illustrates the token transformer module, and the right panel shows the diffusion transformer decoder module. Specifically, given the predicted token sequence, the token transformer first encodes it to a sequential embedding by stacked transformer encoder layers. Then we convert the sequential embedding to a single embedding by applying a max-pooling operation along the temporal dimension. This single embedding is then adopted as condition embedding. For every step of reversed diffusion, we feed the condition embedding, as well as the timestep embedding, and the latent to the diffusion transformer decoder, and estimate the noise $\epsilon_t$. We repeat this reversed diffusion step 1000 times to get the final denoised sample $\textit{X}_0$.\par

\section{More Qualitative Examples}
\setlength{\parindent}{1em}We show more qualitative examples of our method on Text-to-Motion and Audio-to-Motion tasks, respectively. Specifically, we demonstrate the diversity of motion samples generated by our method. Fig.~\ref{fig:supp_t2m_diversity} shows more results on the Text-to-Motion task. In the figure, each column represents 4 samples driven by the same text description. We appropriately adjust the trajectory of some samples for better visualization, so the poses will not clutter together. Fig.~\ref{fig:supp_a2m_diversity} shows more results on the Audio-to-Motion task. Similarly, we show 4 samples driven by the same audio clip in the same column. And we adjust the trajectory of each pose to make them in a two-row formation. As can be observed, our method achieves diversity in both text-driven and audio-driven scenarios, while maintaining semantic correlation. We also provide multiple demonstration videos in our supplementary materials on both text-to-motion and audio-to-motion tasks.\par

\end{document}